\documentclass[11pt]{article}
\usepackage{coling2018}
\usepackage{times}
\usepackage{url}
\usepackage{latexsym}

\usepackage[]{algorithm2e}
\usepackage{algpseudocode}
\usepackage{multirow}
\usepackage{footnote}
\usepackage{xcolor}
\usepackage{colortbl}
\usepackage{amsmath}
\usepackage{amsfonts}
\usepackage[pdftex]{graphicx}
\usepackage{tablefootnote}
\usepackage{fontawesome}

\usepackage{wrapfig}

\usepackage{enumitem}
\usepackage[skip=-1pt]{caption}

\usepackage{fancyvrb}

\usepackage[absolute]{textpos}

\usepackage{tikz}
\newcommand*\circled[1]{\tikz[baseline=(char.base)]{
            \node[shape=circle,draw,inner sep=0.1pt] (char) {#1};}}

\newcommand\scalemath[2]{\scalebox{#1}{\mbox{\ensuremath{\displaystyle #2}}}}

\title{A Practical Incremental Learning Framework \\ For Sparse Entity Extraction}

\author{Hussein S. Al-Olimat$^1$, Steven Gustafson$^2$, Jason Mackay$^3$ \\ {\bf Krishnaprasad Thirunarayan$^1$ and Amit Sheth$^1$} \\
$^1$Kno.e.sis Center, Wright State University, Dayton, OH \\
\texttt{\{hussein;tkprasad;amit\}@knoesis.org} \\
$^2$Maana Inc, Bellevue, WA \\
\texttt{steven.gustafson@maana.io} \\
$^3$GoDaddy Inc, Kirkland, WA \\
\texttt{jmackay@godaddy.com} \\
}

\date{}

\begin{document}

\begin{textblock}{15}(0.5,0.3)
\small
\noindent \textcolor{red}{\textbf{Please cite:}} Hussein S. Al-Olimat, Steven Gustafson, Jason Mackay, Krishnaprasad Thirunarayan, and Amit Sheth. 2018. A practical incremental learning framework for sparse entity extraction. In Proceedings of the 27th International Conference on Computational Linguistics (COLING 2018), pages 700-710. Association for Computational Linguistics. Online: \url{https://www.aclweb.org/anthology/C18-1059/}
\end{textblock}

\maketitle

\vspace{20px}
\begin{abstract}

This work addresses challenges arising from extracting entities from textual data, including the high cost of data annotation, model accuracy, selecting appropriate evaluation criteria, and the overall quality of annotation. We present a framework that integrates Entity Set Expansion (ESE) and Active Learning (AL) methods to reduce the annotation cost of sparse data and provide an online evaluation method as feedback. This incremental and interactive learning framework allows for rapid annotation while maintaining high accuracy.

We evaluate our framework on three publicly available datasets and show that it drastically reduces the cost of sparse entity annotation by an average of 45\% to 85\% while reaching 1.0 and 0.9 F-Score, respectively. Moreover, the method exhibited robust performance across all datasets.

\end{abstract}


\blfootnote{
    \hspace{-0.65cm}  
    This work is licensed under a Creative Commons
    Attribution 4.0 International License.
    License details:
    \url{http://creativecommons.org/licenses/by/4.0/}
}


\section{Introduction}

Entity extraction methods delimit all mentions of an entity class (e.g., Location Names, Proteins, or Auto Parts IDs) in corpora of unstructured text. Supervised machine learning approaches have been effective in extracting such entity mentions, such as \cite{finkel2005incorporating}. However, the prohibitive cost of obtaining a large volume of labeled data for training makes them unattractive and hard to use in realistic settings where resources may be scarce or costly to use (e.g., aviation engineer needing to annotate engine maintenance data). In this paper, we address the problem by developing a practical solution exploiting the strengths of the supervised sequence labeling techniques \cite{ye2009conditional} while reducing the high cost of annotation.

Training data and the annotation approach are the two focal points of any supervised method. A model can be built from pre-annotated data or using \emph{de novo annotations}. There are many pre-annotated and publicly available corpora such as CoNLL-2003 \cite{tjong2003introduction}, GENIA \cite{ohta2001genia}, and BioCreative II \cite{Smith2008}. However, having such data leaves us with two challenges: (i) no training dataset can be found for many of the domain-specific entity classes in enterprise corpora due to data privacy concerns, IP restrictions, and problem specific use cases (needing \emph{de novo annotations}), and (ii) data sparsity, which hinders the performance of supervised models on unseen data (needing \emph{incrementally augmented annotations} to the annotated sentence pool).

We also propose a practical solution to model building
through (i) rapid auto-annotation, to create models with reduced cost,  and (ii) flexible stopping criteria using an online evaluation method, which calculates the confidence of a model on unseen data without a need for a gold standard. Having such online feedback method supports incremental learning which allows us to partially overcome the data sparsity problem (see Section \ref{sec:al}).

Mainstream entity annotation approaches typically present a large body of text or full documents to annotate entities, which can be time-consuming as well as lead to low quality annotations. Additionally, the majority of these techniques require labeling for multiple entity classes which makes the annotation task harder and more complex. Therefore, sentence-level and single-entity-class annotation are desirable to improve tractability and reduce cost. Imposing such requirements, however, causes a labeling starvation problem (i.e., querying annotators to label sentences containing a low frequency of entities from the desired class). In this work, we develop an Entity Set Expansion (ESE) approach to work side by side with Active Learning (AL) to reduce the labeling starvation problem and improve the learning rate.


Our framework is similar to the work proposed by \cite{tsuruoka2008accelerating} that aims to annotate all mentions of a single entity class in corpora while relaxing the requirement of full coverage. Our design choice lowers the annotation cost by using a realistic approach of flexible stopping criteria. It annotates a corpus without requiring annotators to scan all sentences. But it differs from the above in that we use ESE to learn the semantics and retrieve similar entities by analogy to accelerate the learning rate. Moreover, our approach provides Fast (FA), Hyper Fast (HFA), and Ultra Fast (UFA) auto-annotation modes for rapid annotation. Our contributions include:
\begin{enumerate}
 \item A framework to annotate sparse entities rapidly in unstructured corpora using auto-annotation. We then use flexible stopping criteria to learn sequence labeling models incrementally from annotated sentences without compromising the quality of the learned models.
 \item A comprehensive empirical evaluation of the proposed framework by testing it on six entity classes from three public datasets.
 \item An open source implementation of the framework within the widely used Stanford CoreNLP package \cite{manning2014stanford}.\footnote{Codes and data can be found at https://github.com/halolimat/SpExtor}
\end{enumerate}

In the upcoming sections, we describe our incremental learning solution developed using Active Learning, and how we use ESE to accelerate the annotation. Finally, we will evaluate the proposed method on publicly available datasets showing its effectiveness.

\section{Incremental and Interactive Learning}
\label{sec:al}

In a realistic setting, domain experts have datasets from which they want to extract entities, and then use them to build various applications, starting from inverted indexes or catalogs to targeted analysis e.g., to obtain sentiments. Next, we will describe how we use AL to accelerate the learning rate, auto-annotate sentences, and to provide feedback for flexible stopping criteria.

\paragraph{Active Learning (AL)} We adopt the linear-chain Conditional Random Field (CRF) implementation by \cite{finkel2005incorporating} to learn a sequence labeling model to extract entities. However, we use pool-based AL for sequence labeling \cite{settles2008analysis} to replace the sequential or random samplers during online corpus annotation.

Our iterative AL-enabled framework requires a pre-trained/base model to sample the next batch of sentences to be labeled. This iterative sampling allows us to reach higher accuracies with fewer data points by incrementally factoring the new knowledge encoded in the trained models. This more informative sampling is achieved due to the added requirement of model training on all previously annotated sentences as well as new batches of annotated sentences\footnote{We train all CRF models using the default set of features for CoNLL 4 class - https://rebrand.ly/corenlpProp}. In Section \ref{sec:ese}, we show how learning the base model by annotating sentences sampled using a method called Entity Set Expansion (ESE) is better than learning a model from a random sample.

While the default inductive behavior of CRF is to provide one sequence (the most probable one), we instead use an $n$-best sequence method which uses the Viterbi algorithm to return for each sentence the top $n$ sequences with their probabilities. We then query the annotator to annotate $b$ number of sentences (equal to a pre-selected batch size) with the highest entropies \cite{settles2008analysis} calculated using the following equation:



\begin{align}
 nSE(m, s) = - \sum_{\hat{y} \in \hat{Y}} p_m(\hat{y} | s; \theta) \ log \ p_m(\hat{y}|s;\theta)
\label{eq:nse}
\end{align}

\noindent
where $m$ is the trained CRF model, $s$ is a sentence, $\hat{Y}$ is the set of $n$ sequences, and $\theta$ is the features' weights learned by the model $m$.


\paragraph{Annotation Modes}


We devise four AL-enabled annotation modes: {\bf 1.} ESE and AL (EAL), {\bf 2.} Fast (FA), {\bf 3.} Hyper Fast (HFA), and {\bf 4.} Ultra Fast (UFA). The EAL mode (which uses ESE to learn the base model, see Section \ref{sec:ese}) uses model confidence on sentences estimated using $nSE$ for sampling while the rest use a thresholded auto-annotation method that we developed to accelerate the annotation procedure. Auto-annotation employs a pre-trained CRF model $m$ to annotate all unlabeled sentences in the pool. Then, we accept the annotation of sentences from the model if they satisfy the following condition: {\bf if} $\frac{SE_1(s)}{SE_2(s)} \leq t$, where $SE_i(s)$ is the entropy of the sequence $i$ of the sentence $s$ and $t$ is a predefined threshold representing the desired margin difference between the entropies of the two sequences $1$ and $2$. We chose $t$ after running some experiments. What we found was that any threshold below $0.10$ was tight (resulting in no any annotations) and above $0.20$ was large (resulting in many incorrect annotations). Therefore, we chose $0.10$, $0.15$, and $0.20$ for the FA, HFA, and UFA auto-annotation modes, respectively. In Section \ref{sec:fullsyseval}, we evaluate the effectiveness of the annotation modes.


\paragraph{Interactive Learning} We designed an online evaluation method $\sigma$ that provides feedback to annotators on the confidence of a model $m$ on a given sentence pool $S$ (Equation \ref{eq:sigma}). This feedback is an alternative to the F-measure and is very valuable in the absence of a gold standard dataset that could otherwise provide this kind of a feedback.


\newcommand*\mean[1]{\overline{#1}}

\begin{equation}
\mean{nSE(m, S)} = \frac{1}{|S|} \sum_{s \in S}^{} nSE(m, s)
\label{eq:mean_nse}
\end{equation}


\begin{equation}
\sigma = 1-\mean{nSE(m, S)}
\label{eq:sigma}
\end{equation}


\vspace{0.3cm}

$\sigma$ gives the annotator a clearer picture of whether to keep the model as is or learn a new model, interactively, using the mean of all $nSE$s (i.e., $\mean{nSE(m, S)}$). We use $\sigma$ both during the incremental learning steps from the same sentence pool and to test model's accuracy/confidence on unseen sentences. Therefore, using this feedback, we can decide to stop annotating from a certain sentence pool and augment it with sentences from a new pool, or simply stop annotating and train a final model. Consequently, deciding to annotate new sentences that contain novel entities help reduce the effect of data sparsity and increase the accuracy of models, which highlights the importance of this online feedback method.


We compare our online evaluation method with the estimated coverage method in \cite{tsuruoka2008accelerating} that computes the expected number of entities as follows:

\begin{equation}
EC = \frac{E}{E+\sum_{u \in U} E_u}
\label{eq:estimated_coverage}
\end{equation}


\noindent
where $E$ is the number of annotated entities in sentences, $U$ is the set of all unlabeled sentences, and $E_u$ is the expected number of entities in sentence $u$. $E_u$ is calculated by summing the probability of each entity in all of the $n$-best sequences.

\section{Entity Set Expansion Framework}
\label{sec:ese}

Since AL requires a base (pre-trained) model to work \cite{settles2008analysis}, learning that model from annotated sentences sampled using sequential or random sampling can be very expensive due to the labeling starvation problem mentioned before. Therefore, we developed an Entity Set Expansion (ESE) method that incorporates and exploits the semantics of the desired entity class to more informatively sample sentences that are likely to have entities of the desired class.

We assume that all noun phrases (NPs) are candidate entities and extract all of them from the unstructured text.\footnote{While not all candidate NPs are positive examples of an entity, a human-in-the-loop would interactively give feedback to the system by choosing the positive ones.} Then, we record five features for each noun phrase ($np$) and model the data as a bipartite graph with NPs on one side and features on the opposite side, which we describe next.\footnote{The bipartite graph is used in order to allow for calculating the similarity between noun phrases by modeling the edges from multi-modal edge weights.}

\subsection{Noun Phrase Extraction (NPEx)}
\label{sec:npex}


The first step of our method is to extract NPs. We POS tag and parse sentences to the form: $[w_i/pos_i \ \forall w \in W]$, then, we use the following regular expressions to extract the NPs:

{\small
\begin{verbatim}
      NP = JJs + NNs + CDs
      JJs = (?:(?:[A-Z]\\w+ JJ )*)
      NNs = (?:[^\\s]* (?:N[A-Z]*)\\s*)+
      CDs = (?:\\w+ CD)?
\end{verbatim}
}

\subsection{Featurization}
\label{sec:featurization}

We automatically extract and record five kinds of features for each noun phrase that we extract from the text (see Section \ref{sec:npex}). The list of features follows:

\begin{enumerate}[label=\protect\circled{\arabic*},leftmargin=1cm]
\item \textbf{Lexical Features (LF):}
    \begin{itemize}
        \item {\bf Orthographic Form (OF)}: We abstract OF features from the actual word and obtain its type (i.e., numeric, alpha, alphanumeric, or other). Additionally, we classify the word as: all upper, all lower, title case, or mixed case.
        \item {\bf Word Shape (WS)}: We define WS to abstract the patterns of letters in a word as a short/long word shape (SWS/LWS) features. In LWS, we map each letter to ``L'', each digit to ``D'', and retain the others. On the other hand, for SWS, we remove consecutive character types. For example, LWS(ABC-123) $\rightarrow$ ``LLL-DDD'' and SWS(ABC-123) $\rightarrow$ ``L-D''.
    \end{itemize}
\item \textbf{Lexico-Syntactic Features (LS):}
    We use a skip-gram method to record the explicit LS-patterns surrounding each NP, i.e., for each $np$ in $s$ we record the pattern $w_{i-1}+np+w_{i+1}$, where $w_{(.)}$ are the two words that precede and follow $np$.
\item \textbf{Syntactic Features (SF):}
    We use dependency patterns to abstract away from the word-order information that we can capture using the contextual and lexico-syntactic features. We use Stanford's English\_UD neural network-based dependency parser \cite{chen2014fast} to extract universal dependencies of NPs. For each $np$ in $s$, we record two dependency patterns of the NP that serve in \emph{governor} and \emph{dependent} roles.

\item \textbf{Semantic Features (SeF):}
    To capture the lexical semantics, we use WordNet \cite{miller1995wordnet} to get word senses and draw sense relations between NPs if they have the same sense class.
\item \textbf{Contextual Features (CF):}
    To capture the latent features, we use a Word2Vec embedding model \cite{mikolov2013distributed} trained on the sentence pool $S$ as the only bottom-up distributional semantics method. We exploit the word-context co-occurrence patterns learned by the model to induce the relational similarities between NPs.
\end{enumerate}

The use of semantic (i.e., word senses) and contextual (i.e., word embeddings) features on sparse or domain-specific data (such as enterprise data) is not enough and sometimes not even possible due to unavailability \cite{tao2015leveraging}. Therefore, we tried to use diverse features in a complementary manner to capture as many meaningful relations between potential entities as possible. For example, the absence of SeF for domain specific entities, such as protein molecules or parts numbers, is compensated by the use of WS-LF (e.g., by drawing a relation between the NPs ``IL-2'' and ``AP-1'').


\subsection{Feature-Graph Embedding}
\label{sec:graph_embedding}

We model edges in the bipartite graph by assigning a weight $w$ between each pair of a noun phrase $n$ and a feature $f$ using one of the following:


\begin{equation}
w^1_{n,f} = C_{n,f}
\label{eq:count}
\end{equation}

\begin{equation}
w^2_{n,f} = log (1+C_{n,f})[log|N|-log(|N|_f)]
\label{eq:tfidf}
\end{equation}

\begin{equation}
w^3_{n,f} = log (1+C_{n,f})[log|N|-log(\sum_{\hat{n}} C_{\hat{n},f})]
\label{eq:tfidf_sum}
\end{equation}

\noindent
where 
$C_{n,f}$ is the co-occurrence count between $n$ and $f$, $|N|$ is the number of NPs in our dataset, $|N|_f$ is the co-occurrence count of all NPs with $f$, and $\sum_{\hat{n}} C_{\hat{n},f}$ is the sum of all NP co-occurrences with the feature $f$. Equations \ref{eq:tfidf} and \ref{eq:tfidf_sum} are two variations of TFIDF and the latter is adapted to weigh the edges in \cite{shensetexpan,rong2016egoset}.


\setlength{\columnsep}{15pt}
\begin{wrapfigure}[16]{r}{0.5\textwidth}
\vspace{-0.3cm}
\begin{minipage}{0.5\textwidth}
\begin{algorithm}[H]
 \KwData{$e$: input seed; $S$: text sentences}
 \KwResult{$N = \{n\}$: ranked similar noun phrases}
 \bf Start \\

 $\hat{N}$ = $\emptyset$; \tcp{all noun phrases}
 $F$ = $\emptyset$; \tcp{all features}

 \For{$s \gets S$}{
    $\hat{N} =$ $\hat{N}$ $\cup$ ExtractNounPhrases($s$); \\
    $F = F$ $\cup$ Featurize($\hat{N}$); {\tcp{Section \ref{sec:featurization}}}
 }
 $G =$ BuildBipartiteGraph($\hat{N}$, $F$); {\tcp{Section \ref{sec:graph_embedding}}}

 $N =$ CalculateSimilarity($e$, $G$); {\tcp{Equation \ref{eq:cosinesim} or \ref{eq:contextdepsim}}}
 \Return $rank(N)$; \\
 \caption{Entity Set Expansion}
 \label{alg:ese}
\end{algorithm}
\end{minipage}
\end{wrapfigure}


\subsection{Set Expansion}
\label{sec:setexpan}

\begin{figure*}
    \includegraphics[width=0.8\textwidth]{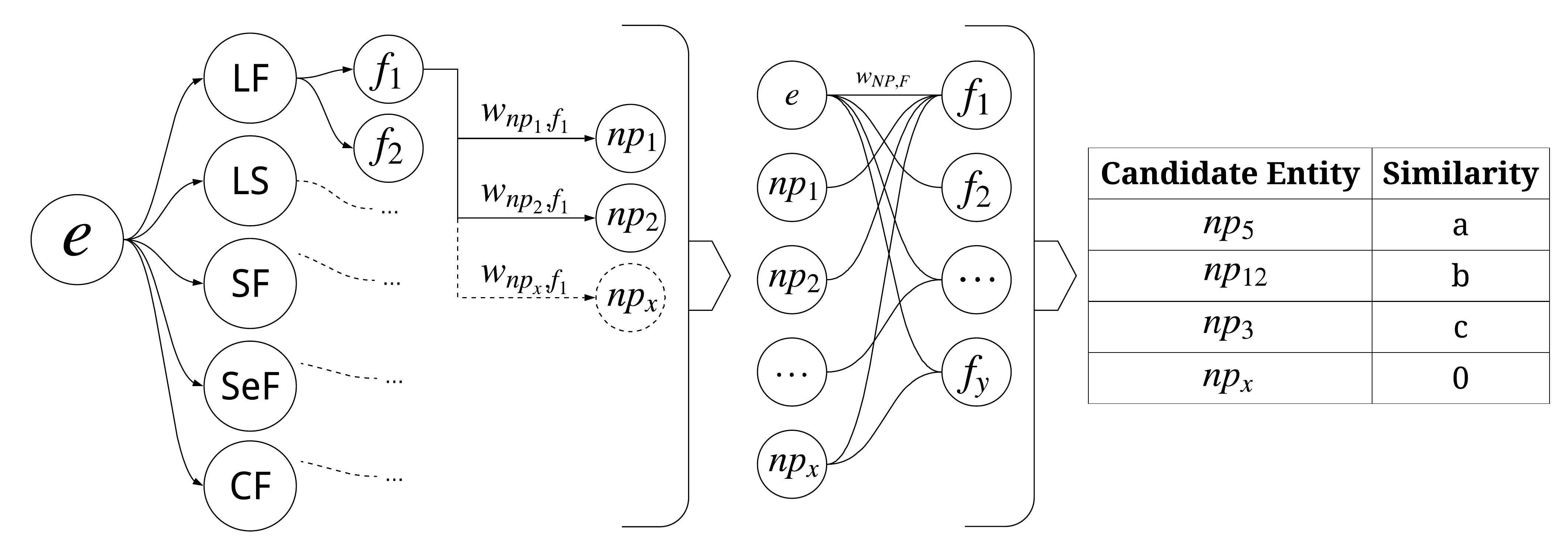}
    \centering
    \caption{The three stages of graph embedding for candidate entities ranking.}
    \label{fig:f_graph_embedding}
\end{figure*}

Our method starts by taking a seed entity $e$ from the annotator input, and returns a ranked list of similar NPs (See Algorithm \ref{alg:ese}). After constructing a graph using the embedding method mentioned in the previous section (See Fig. \ref{fig:f_graph_embedding}), we calculate the similarity between the seed entity $e$ and all other noun phrases $NPs$ in the graph $G$ using one of the following similarity methods:

\begin{equation}
 Sim^1(n_1, n_2 | F) = \scalemath{0.9}{\frac{\sum\limits_{f \in F} w_{n_1,f} w_{n_2,f}}{\sqrt{\sum\limits_{f \in F} w_{n_1,f}^2}\sqrt{\sum\limits_{f \in F} w_{n_2,f}^2}}}
 \label{eq:cosinesim}
\end{equation}

\begin{equation}
Sim^2(n_1, n_2 | F) = \scalemath{0.9}{\frac{\sum\limits_{f \in F} \min(w_{n_1,f}, w_{n_2,f})}{\sum\limits_{f \in F}\max(w_{n_1,f}, w_{n_2,f})}}
\label{eq:contextdepsim}
\end{equation}


\noindent
where $Sim(n_1, n_2 | F)$ is the similarity between the two noun phrases $n_1$ and $n_2$ given the set of features $F$ they have in common, and $w_{(.,.)}$ is the weight of the edge between the NPs and features (defined using one of the Equations \ref{eq:count}-\ref{eq:tfidf_sum}). Equation \ref{eq:cosinesim} is the cosine similarity and Equation \ref{eq:contextdepsim} is the context-dependent similarity by \cite{shensetexpan}.


\subsection{Feature Ensemble Ranking}
\label{sec:ensembling}

\begin{figure*}
    \includegraphics[width=0.8\textwidth]{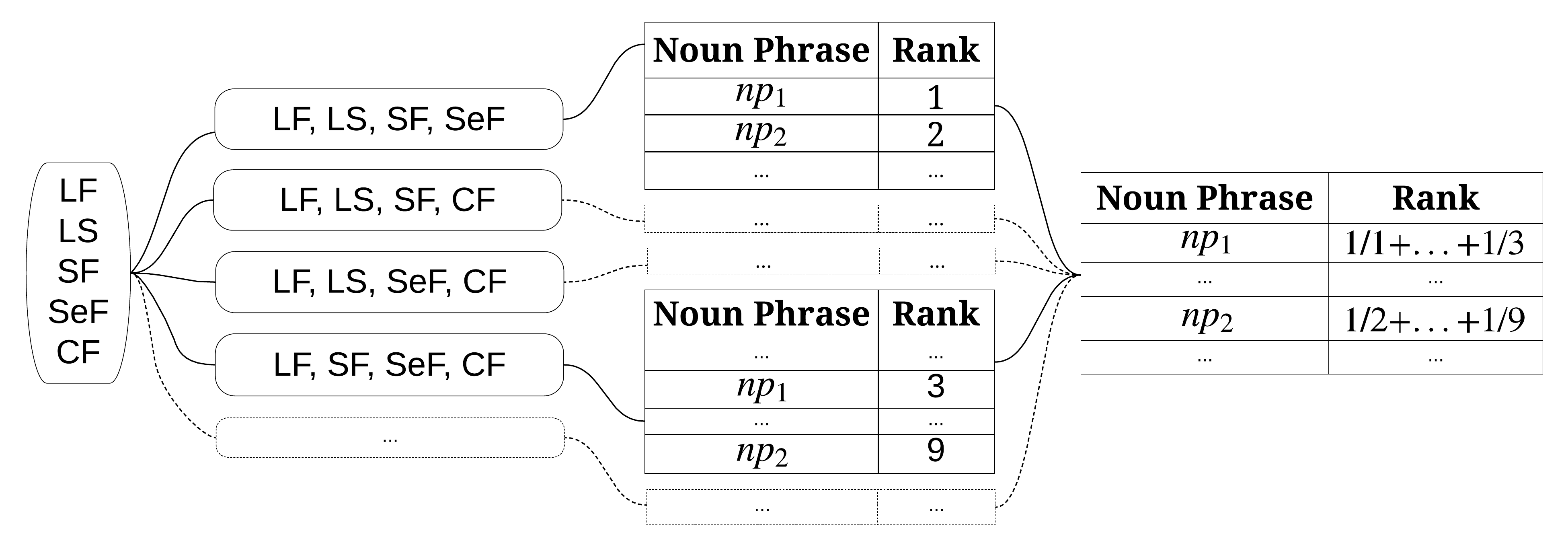}
    \centering
    \caption{Coarse features ensemble method for ranking candidate entities.}
    \label{fig:f_ensemble}
\end{figure*}

Similar to the use in SetExpan \cite{shensetexpan}, we ensemble the coarse features\footnote{Coarse features are the five groups of features in Section \ref{sec:featurization} as opposed to the fine ones which are part of each of them.} and rank NPs in sublists to reduce the effect of inferior features on the final ranking. The size of each sublist is equal to $|\hat{F}|-1$, where $\hat{F}$ is the set of all features from Section \ref{sec:featurization}. We then use the mean reciprocal rank (Equation \ref{eq:mrr}) to find the final ranking of each NP using the rank from each sublist (See Fig. \ref{fig:f_ensemble}).

\begin{equation}
MRR = \frac{1}{Q} \sum_{i=0}^{|Q|} \frac{1}{rank_i}
\label{eq:mrr}
\end{equation}



\section{Experiments and Results}

\subsection{Data Preparation}

\setlength{\tabcolsep}{0.23em} 
{\renewcommand{\arraystretch}{1.4}
\begin{table*}[t!]
\small
\begin{center}
\begin{tabular}{|l|l|l|l|l|l|}
\hline

\bf Dataset Name & \bf Entity Class & \bf $|S|$ (with Entities) & \bf \# Entities & \bf NPEx Accuracy & \bf Seed Entity (Count) \\ \hline
\multirow{3}{*}{CoNLL-2003} & Location & 13519 (36\%) & 1258 & 84\% & U.S. (296), Washington (26)\\ \cline{2-6}
 & Person & 13519 (31\%) & 3388 & 76\% & Clinton (70), Von Heesen (1) \\ \hline
 BioCreAtIvE II & Gene & 12500 (51\%) & 10441 & 51\% & insulin (64), GrpE (1) \\ \hline
\multirow{3}{*}{GENIA 3.02} & Protein Molecule & 14838 (55\%) & 3413 & 60\% & NF-kappa B (552), IL-1RA (1) \\ \cline{2-6}
 & Cell Type & 14838 (27\%) & 1569 & 24\% & T cells (489), MGC (2) \\ \cline{2-6}
 & Virus & 14838 (8\%) & 324 & 50\% & HIV-1 (336), adenovirus E1A (1) \\ \hline

\end{tabular}
\end{center}
\caption{Evaluation datasets statistics, noun phrase extraction accuracies, and seed entities counts.}
\label{tbl:datasets}
\end{table*}

To test our framework while emulating the full annotation experience of annotators, we use three publicly available gold standard datasets (CoNLL-2003, GENIA 3.02, and BioCreAtIvE II) labeled for several entity classes.

We tokenize documents into sentences, then query the emulator (which plays the role of annotators) to label sentences, thus avoiding the required annotation of full documents for better user experience. We require the emulator to label only a single entity class from each dataset, to avoid the complexity of multiple annotations. Therefore, we created six versions of the datasets where only one class is kept in each version. Table \ref{tbl:datasets} includes some statistics of the datasets. The percentage of sentences with entities of a given class shows the sparsity of those entities.


\subsection{NPEx Evaluation}


%
%
%

Since ESE method operates at an NP level, the performance of Noun Phrase Extraction (NPEx) influences the overall performance of ESE significantly. Table \ref{tbl:datasets} includes the accuracy of NPEx as the percentage of the actual class entities that were extracted as candidate entities (NPs). In the future, ESE performance can be improved by enhancing the performance of candidate entities extraction through, for example, extracting noun phrases formed from typed-dependencies.


\subsection{Entity Set Expansion (ESE) Evaluation}

%
%
%
%
%

To test the influence of seed entity frequency on ESE performance, we manually picked two seeds, the most frequent and the least frequent among all noun phrases (See Table \ref{tbl:datasets}). Additionally, we varied the weighting measure of the graph edges using one of the three equations (\ref{eq:count}-\ref{eq:tfidf_sum}). Finally, we varied the similarity measure when ranking NPs (Equations \ref{eq:cosinesim} and \ref{eq:contextdepsim}).

\begin{table*}[t!]
\small
\begin{center}
\begin{tabular}{cc|c|c|c|c|c|c|c|c|c|c|c|c|c|c|c|c|c|c|}
\cline{3-20}

& &
\multicolumn{3}{c|}{ \bf Location } &
\multicolumn{3}{c|}{ \bf Person } &
\multicolumn{3}{c|}{ \bf Gene } &

\multicolumn{3}{c|}{ \bf Protein } &
\multicolumn{3}{c|}{ \bf Cell Type } &
\multicolumn{3}{c|}{ \bf Virus } \\ \cline{3-20}

& &
Eq.\ref{eq:count} & Eq.\ref{eq:tfidf} & Eq.\ref{eq:tfidf_sum} &
Eq.\ref{eq:count} & Eq.\ref{eq:tfidf} & Eq.\ref{eq:tfidf_sum} &
Eq.\ref{eq:count} & Eq.\ref{eq:tfidf} & Eq.\ref{eq:tfidf_sum} &
Eq.\ref{eq:count} & Eq.\ref{eq:tfidf} & Eq.\ref{eq:tfidf_sum} &
Eq.\ref{eq:count} & Eq.\ref{eq:tfidf} & Eq.\ref{eq:tfidf_sum} &
Eq.\ref{eq:count} & Eq.\ref{eq:tfidf} & Eq.\ref{eq:tfidf_sum} \\ \hline

\multicolumn{1}{|c}{}\multirow{2}{*}{\bf Seed 1} & \multicolumn{1}{|c|}{Eq.\ref{eq:cosinesim}} &
0.37 & 0.40 & 0.50 & 0.23 & 0.23 & 0.30 & 0.00 & 0.03 & 0.13 & 0.17 & 0.23 & 0.20 & 0.27 & 0.50 & 0.53 & 0.20 & 0.13 & 0.17  \\ \cline{2-20}

\multicolumn{1}{|c}{} & \multicolumn{1}{|c|}{Eq.\ref{eq:contextdepsim}} &
0.63 & \bf 0.73 & 0.73 & 0.03 & \bf 0.17 & 0.20 & 0.03 & \bf 0.07 & 0.07 & 0.43 & \bf 0.43 & 0.53 & 0.17 & \bf 0.23 & 0.23 & 0.07 & \bf 0.10 & 0.07 \\ \hline

\multicolumn{1}{|c}{}\multirow{2}{*}{\bf Seed 2} & \multicolumn{1}{|c|}{Eq.\ref{eq:cosinesim}} &
0.33 & 0.33 & 0.57 & 0.53 & 0.40 & 0.30 & 0.63 & 0.63 & 0.63 & 0.17 & 0.60 & 0.27 & 0.10 & 0.20 & 0.13 & 0.07 & 0.03 & 0.03  \\ \cline{2-20}

\multicolumn{1}{|c}{} & \multicolumn{1}{|c|}{Eq.\ref{eq:contextdepsim}} &
0.57 & \bf 0.70 & 0.63 & 0.47 & \bf 0.37 & 0.37 & 0.60 & \bf 0.57 & 0.60 & 0.07 & \bf 0.30 & 0.30 & 0.07 & \bf 0.07 & 0.07 & 0.03 & \bf 0.10 & 0.03 \\

\hline

\end{tabular}
\end{center}
\caption{ESE performance ($p@k$). Best performing combination is bold faced.}
\label{tbl:eseperformance}
\end{table*}

We tested the performance of ESE with and without using the feature ensemble method in Section \ref{sec:ensembling}. We measured the precision of the method in ranking positive examples of entities similar to the seed entity (Seed 1 and Seed 2) in the top $k$ NPs. We designed ESE to output thirty candidate entities (NPs) ranked based on the similarity to the seed term. Therefore, we calculate precision at $k$ ($p@k$) where $k$ is always 30. Table \ref{tbl:eseperformance} shows the best results when using the feature ensemble method which is more stable than the non-ensemble one (due to lower standard deviation and non-zero precisions). According to the results, the best combination in terms of the mean and standard deviation is obtained when using TFIDF (Eq.\ref{eq:tfidf}) to weigh the edges and the context-dependent similarity (Eq.\ref{eq:contextdepsim}) to rank NPs.


\subsection{Full System Evaluation}
\label{sec:fullsyseval}

\begin{figure}
    \includegraphics[width=0.6\textwidth]{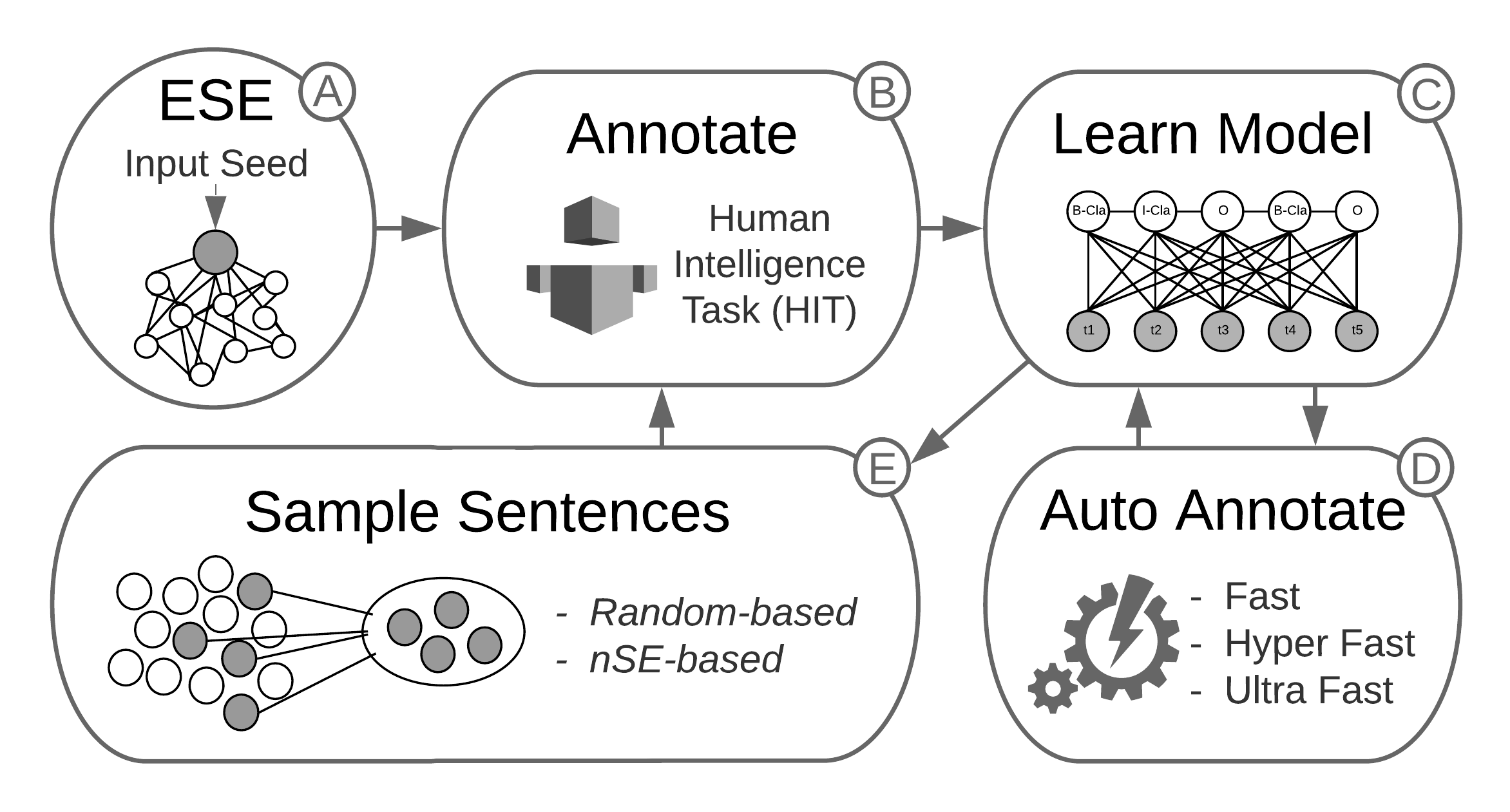}
    \centering
    \caption{End-to-End system pipeline. Arrows represent paths can be followed to annotate text.}
    \label{fig:frameworkFlow}
\end{figure}



We tested our pipeline on three different settings following three different paths in Fig. \ref{fig:frameworkFlow}:
\begin{enumerate}
 \item All Random (AR): $(E, B, C)^*$, where $E$ is a random-based sampler.
 \item ESE and AL (EAL): $A, B, C, (E, B, C)^*$, where $E$ is an $nSE$-based sampler.
 \item EAL in addition to auto-annotation (EAA): $A, B, C, (D, C, E, B, C)^*$
\end{enumerate}

\noindent
We iterate through the loop paths in the starred parentheses while sampling 100 sentences each time until we finish all sentences in the pool or reach full F-Score. In Fig. \ref{fig:learningCurves}, we show the performance of the first two settings (i.e., AR and EAL) in terms of F-Score. The use of AL and ESE methods outperformed random sampling all the time. ESE increased the performance of the base model by 35\% F-Score on average, allowing us to reach 0.5 F-Score while the random sampler reached 0.37 F-Score only.


\setlength{\columnsep}{15pt}
\begin{figure}
\vspace{-0.2cm}
    \includegraphics[width=0.7\textwidth]{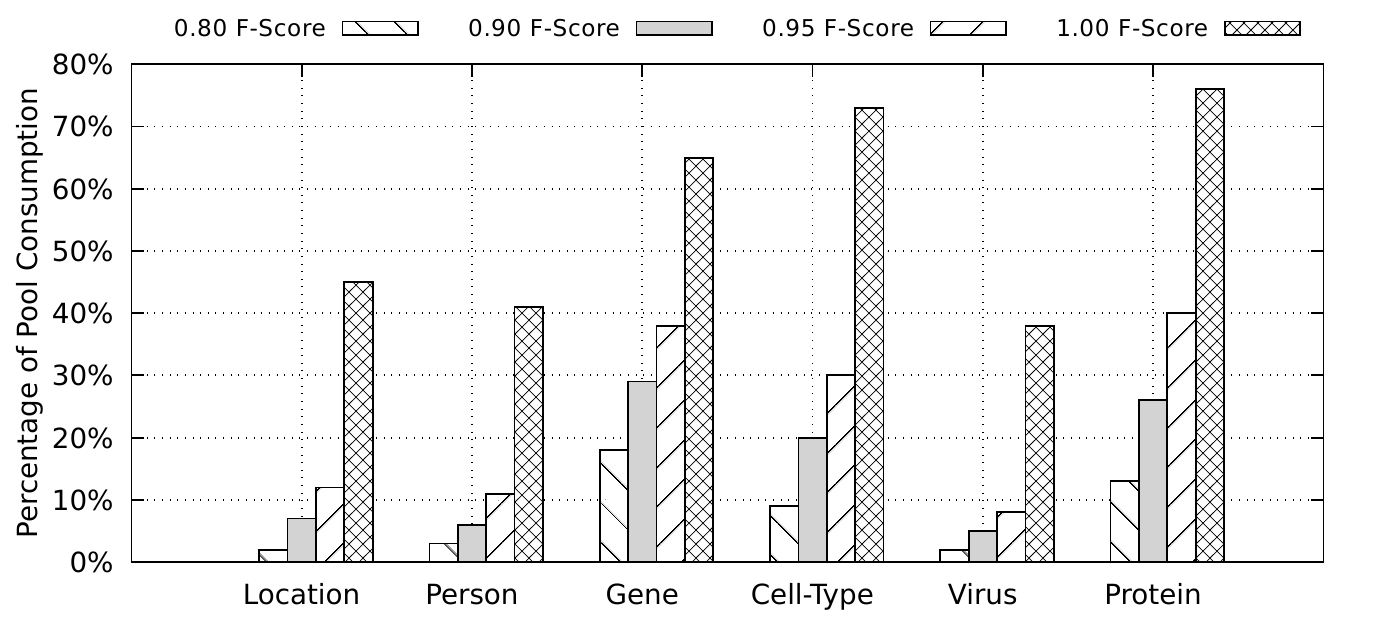}
    \centering
    \caption{Percentage of sentences annotated while using EAL to reach different F-Scores.}
    \label{fig:percentageConsumption}
\end{figure}

As shown in Fig. \ref{fig:percentageConsumption}, the percentage consumption of the sentence pool to reach up to 0.95 F-Score follows almost a linear growth. However, this cost grows exponentially if we want to reach 1.0 F-Score (needing on average around 33\% more sentences). Practically speaking, we might want to sacrifice the 5\% F-Score improvement to significantly minimize manual labor.\footnote{This would, therefore, make the annotated data a silver standard instead of a gold standard.}


\begin{table*}[t!]
\small
\begin{center}
\begin{tabular}{|l|l|l|l|l|l|l|}

\hline

\multicolumn{1}{|c|}{\bf \multirow{3}{*}{Dataset Name}} & \multicolumn{1}{c|}{\bf \multirow{3}{*}{Entity Class}} & \multicolumn{2}{c|}{\bf EAL} & \multicolumn{3}{c|}{\bf EAA Annotation Mode} \\ \cline{3-7}
 & & \multicolumn{2}{c|}{@ 1.0 F} & \bf FA & \bf HFA & \bf UFA \\ \cline{3-7}
 & & $\sigma$ & \% cut & \multicolumn{3}{c|}{\bf F-Score (percentage cut)} \\ \hline

\multirow{3}{*}{CoNLL-2003} & Location & 0.97 & 55\% & 0.99 (46\%) & 0.93 (83\%) & 0.82 (91\%) \\ \cline{2-7}
                            & Person & 0.97 & 59\% & 0.99 (48\%) & 0.95 (81\%) & 0.85 (90\%) \\ \hline
BioCreAtIvE II & Gene & 0.94 & 35\% & 1.00 (35\%) & 0.96 (50\%) & 0.89 (69\%) \\ \hline

\multirow{3}{*}{GENIA 3.02} & Protein Molecule & 0.99 & 33\% & 0.98 (36\%) & 0.87 (71\%) & 0.74 (85\%) \\ \cline{2-7}
                            & Cell Type & 0.99 & 62\% & 0.94 (70\%) & 0.82 (86\%) & 0.74 (91\%) \\ \cline{2-7}
                            & Virus & 0.94 & 24\% & 0.97 (79\%) & 0.89 (94\%) & 0.84 (96\%) \\ \hline

\multicolumn{2}{|c|}{\bf Average} & 0.97 & 45\% & 0.98 (52\%) & 0.90 (78\%) & 0.81 (87\%) \\ \hline

\end{tabular}
\end{center}
\caption{Pipeline testing results of EAL and EAA annotation modes showing the model confidence ($\sigma$), F-Scores, and percentage cut from the pool of sentences.}
\label{tbl:autoannotation}
\end{table*}

We used Jensen-Shannon divergence \cite{lin1991divergence} to compare the curves corresponding to the performance of the estimated coverage method (Equation \ref{eq:estimated_coverage}) and our online evaluation metric $\sigma$ (Equation \ref{eq:sigma}) with the F-Score curve of EAL (see Fig. \ref{fig:learningCurves}). Our method $\sigma$ outperformed the estimated coverage method with a decrease of 96\% in dissimilarity, which makes our method more reliable. Additionally, as shown in Table \ref{tbl:autoannotation}, on average $\sigma$ gives an accurate estimation of the F-Score without labeled data with an error margin of 3\% on average when reaching 1.0 F-Score (i.e., @ 1.0 F).\footnote{The $\sigma$ curves of GENIA-Cell-Type and GENIA-Virus shows high overestimation of the F-Score curve in the first iterations. This is due to missing entity annotations we found in the gold standard, which mistakenly shows that we had many false positives. To list a few, GENIA-Cell-Type have missing annotations for ``transformed T cells'', ``transformed cells'', and ``T cells'', where GENIA-Virus have some missing annotations for ``HIV-1'' and ``HIV-2''.}

\begin{figure}[ht]
    \includegraphics[width=1\textwidth]{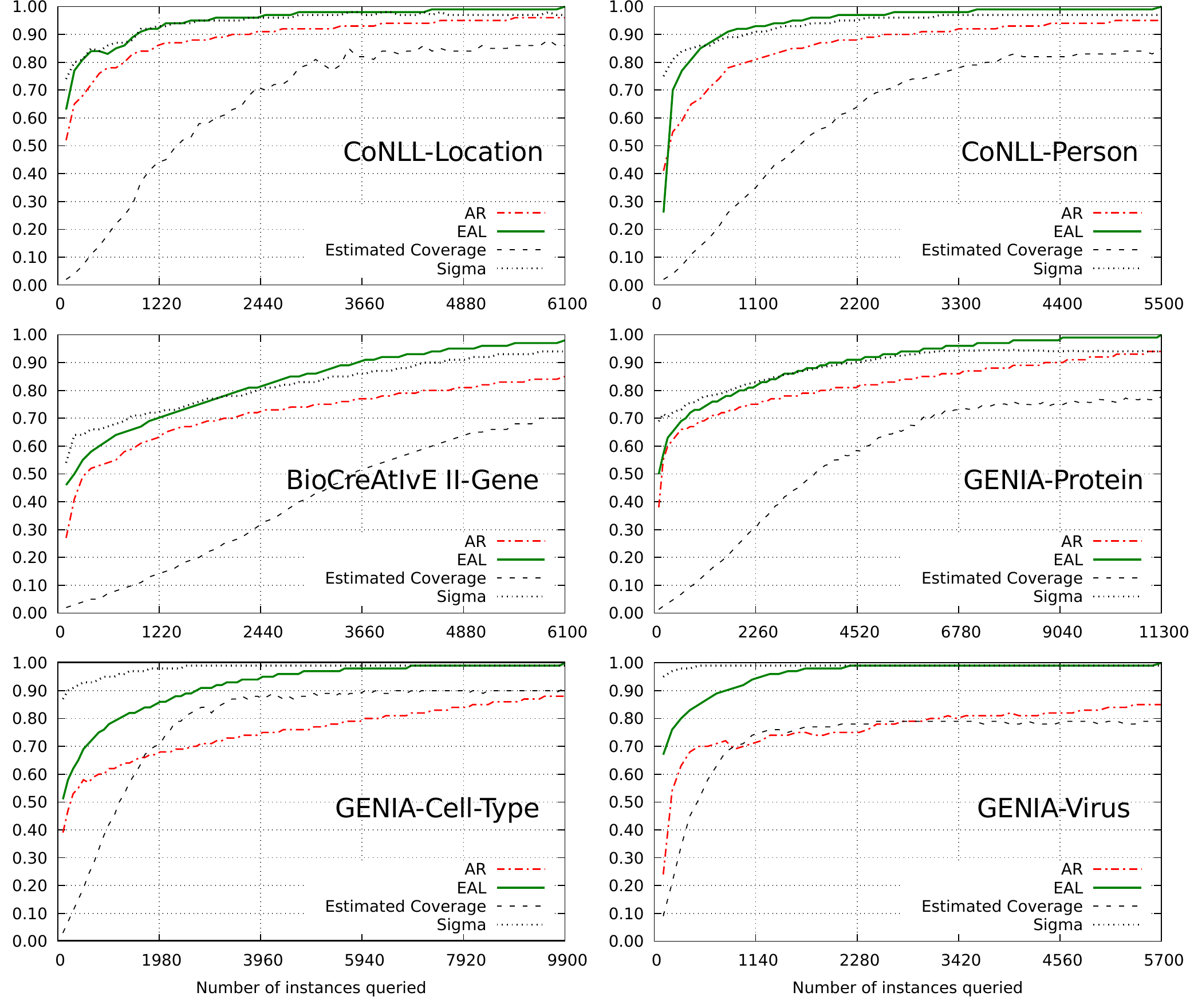}
    \centering
    \caption{Learning curves with different query strategy. Y-axis value for AR and EAL is the F-Score, and for Sigma and Estimated Coverage is the value from Equations \ref{eq:sigma} and \ref{eq:estimated_coverage}, respectively.}
    \label{fig:learningCurves}
\end{figure}

Our EAL method also outperformed \cite{tsuruoka2008accelerating} on the three overlapping entity classes we tested on from the two datasets (Genia and CoNLL-2003). EAL needed 2800, 2900, and 400 fewer sentences to reach the same coverage as \cite{tsuruoka2008accelerating} for the Location, Person, and Cell-Type datasets, respectively.

Finally, for the last setting, we tested the system using the three auto-annotation modes (i.e., FA, HFA, and UFA) as shown in Table \ref{tbl:autoannotation}. While the auto-annotation mode can allow us to reduce up to 87\% of the data pool, this drastic saving also reduces the accuracy of the learned model, achieving, on average, around 81\% F-Score. Overall, our framework presents a trade off between coverage and annotations cost. The HFA auto-annotation mode shows the benefit, especially in a realistic enterprise setting, where the cost to annotate 33\% of the data to only increase 10\% F-Score (when comparing the on average performance of HFA with ESA) is unreasonable.

Table \ref{tbl:autoannotation} appears to show FA being inferior to EAL for the Location class, for example. In reality FA reduced sentence annotation by 65\% to reach 0.99 F-Score. Further, as our testing criteria demanded that we either reach 1.0 F-Score or finish the sentences, FA tried to finish the pool without performance improvement.




\section{Related Work}

To extract sparse entities from texts and learn sequence models faster, our practical framework use Active Learning (AL) for sequence labeling \cite{tomanek2009semi,settles2008analysis}, both as a method of sampling and to auto-annotate sentences. Our work encompasses entity extraction, Entity Set Expansion (ESE), corpora pre-annotation, auto-annotation, rapid corpora annotation, and AL.


Using a pattern-based ESE (a.k.a., seed set expansion) technique on top of AL helped our approach in discovering rare patterns and rules which might have been hidden when using only a feature-based system \cite{chiticariu2013rule}. Our ESE method is similar to \cite{shensetexpan,tao2015leveraging,sadamitsu2012entity} where the system starts with a few positive examples of an entity class and tries to extract similar entities. Additionally, we use a richer set of features than those in \cite{gupta2014improved,grishman2014information,min2011fine} while training a CRF model and using it to further find positive examples of a given entity class.

Methods such as \cite{sarmento2007more} computes the degree of membership of an entity to a group of predefined entities called seed entities where the class of each group is predefined. Their method does not focus on entity extraction. Rather it focuses on detecting the class of entities during the evaluation step, which makes their method different from what we are trying to achieve here.




Regarding corpus annotation, many notable previous works such as \cite{lingren2013evaluating} use dictionaries to pre-annotate texts. However, inaccurate pre-annotations may harm more than improve since they require an added overhead of modifications and deletions \cite{rehbein2009assessing}. \newcite{kholghi2017active} and \newcite{tsuruoka2008accelerating} propose AL-based annotation systems. However, their work differs from ours in the following ways: 
{\bf (1)} we propose a more accurate online evaluation method than theirs, {\bf (2)} we use ESE to bootstrap the learning framework collaboratively with a user-in-the-loop, and finally, {\bf (3)} we provide auto-annotation modes which reduces the number of sentences to be considered for annotation and so allows for better usability of the framework.

\section{Conclusions and Future Work}

We presented a practical and effective solution to the problem of sparse entity extraction. Our framework builds supervised models and extracts entities with a reduced annotation cost using Entity Set Expansion (ESE), Active Learning (AL), and auto-annotation without compromising extraction quality. Additionally, we provided an online method for evaluating model confidence that enables flexible stopping criteria. In the future, we might consider evaluating different AL querying strategies and compare their performances.



\bibliography{references}

\begin{thebibliography}{}

\bibitem[\protect\citename{Chen and Manning}2014]{chen2014fast}
Danqi Chen and Christopher Manning.
\newblock 2014.
\newblock A fast and accurate dependency parser using neural networks.
\newblock In {\em Proceedings of the 2014 conference on empirical methods in
  natural language processing (EMNLP)}, pages 740--750.

\bibitem[\protect\citename{Chiticariu \bgroup et al.\egroup
  }2013]{chiticariu2013rule}
Laura Chiticariu, Yunyao Li, and Frederick~R Reiss.
\newblock 2013.
\newblock Rule-based information extraction is dead! long live rule-based
  information extraction systems!
\newblock In {\em EMNLP}, number October, pages 827--832.

\bibitem[\protect\citename{et. al.}2008]{Smith2008}
Larry~Smith et. al.
\newblock 2008.
\newblock Overview of biocreative ii gene mention recognition.
\newblock {\em Genome Biology}, 9(2):S2, Sep.

\bibitem[\protect\citename{Finkel \bgroup et al.\egroup
  }2005]{finkel2005incorporating}
Jenny~Rose Finkel, Trond Grenager, and Christopher Manning.
\newblock 2005.
\newblock Incorporating non-local information into information extraction
  systems by gibbs sampling.
\newblock In {\em Proceedings of the 43rd annual meeting on association for
  computational linguistics}, pages 363--370. Association for Computational
  Linguistics.

\bibitem[\protect\citename{Grishman and He}2014]{grishman2014information}
Ralph Grishman and Yifan He.
\newblock 2014.
\newblock An information extraction customizer.
\newblock In Petr Sojka, Ale{\v{s}} Hor{\'a}k, Ivan Kope{\v{c}}ek, and Karel
  Pala, editors, {\em Text, Speech and Dialogue}, pages 3--10, Cham. Springer
  International Publishing.

\bibitem[\protect\citename{Gupta and Manning}2014]{gupta2014improved}
Sonal Gupta and Christopher~D Manning.
\newblock 2014.
\newblock Improved pattern learning for bootstrapped entity extraction.
\newblock In {\em CoNLL}, pages 98--108.

\bibitem[\protect\citename{Kholghi \bgroup et al.\egroup
  }2017]{kholghi2017active}
Mahnoosh Kholghi, Laurianne Sitbon, Guido Zuccon, and Anthony Nguyen.
\newblock 2017.
\newblock Active learning reduces annotation time for clinical concept
  extraction.
\newblock {\em International journal of medical informatics}, 106:25--31.

\bibitem[\protect\citename{Kim \bgroup et al.\egroup }2006]{kim2006mmr}
Seokhwan Kim, Yu~Song, Kyungduk Kim, Jeong-Won Cha, and Gary~Geunbae Lee.
\newblock 2006.
\newblock Mmr-based active machine learning for bio named entity recognition.
\newblock In {\em Proceedings of the Human Language Technology Conference of
  the NAACL, Companion Volume: Short Papers}, pages 69--72. Association for
  Computational Linguistics.

\bibitem[\protect\citename{Lin}1991]{lin1991divergence}
Jianhua Lin.
\newblock 1991.
\newblock Divergence measures based on the shannon entropy.
\newblock {\em IEEE Transactions on Information theory}, 37(1):145--151.

\bibitem[\protect\citename{Lingren \bgroup et al.\egroup
  }2013]{lingren2013evaluating}
Todd Lingren, Louise Deleger, Katalin Molnar, Haijun Zhai, Jareen Meinzen-Derr,
  Megan Kaiser, Laura Stoutenborough, Qi~Li, and Imre Solti.
\newblock 2013.
\newblock Evaluating the impact of pre-annotation on annotation speed and
  potential bias: natural language processing gold standard development for
  clinical named entity recognition in clinical trial announcements.
\newblock {\em Journal of the American Medical Informatics Association},
  21(3):406--413.

\bibitem[\protect\citename{Manning \bgroup et al.\egroup
  }2014]{manning2014stanford}
Christopher~D Manning, Mihai Surdeanu, John Bauer, Jenny~Rose Finkel, Steven
  Bethard, and David McClosky.
\newblock 2014.
\newblock The stanford corenlp natural language processing toolkit.
\newblock In {\em ACL (System Demonstrations)}, pages 55--60.

\bibitem[\protect\citename{Mikolov \bgroup et al.\egroup
  }2013]{mikolov2013distributed}
Tomas Mikolov, Ilya Sutskever, Kai Chen, Greg~S Corrado, and Jeff Dean.
\newblock 2013.
\newblock Distributed representations of words and phrases and their
  compositionality.
\newblock In {\em Advances in neural information processing systems}, pages
  3111--3119.

\bibitem[\protect\citename{Miller}1995]{miller1995wordnet}
George~A Miller.
\newblock 1995.
\newblock Wordnet: a lexical database for english.
\newblock {\em Communications of the ACM}, 38(11):39--41.

\bibitem[\protect\citename{Min and Grishman}2011]{min2011fine}
Bonan Min and Ralph Grishman.
\newblock 2011.
\newblock Fine-grained entity set refinement with user feedback.
\newblock {\em Information Extraction and Knowledge Acquisition}, page~2.

\bibitem[\protect\citename{Ohta \bgroup et al.\egroup }2001]{ohta2001genia}
Tomoko Ohta, Yuka Tateisi, Jin-Dong Kim, Sang-Zoo Lee, and Jun’ichi Tsujii.
\newblock 2001.
\newblock Genia corpus: A semantically annotated corpus in molecular biology
  domain.
\newblock In {\em Proceedings of the ninth International Conference on
  Intelligent Systems for Molecular Biology (ISMB 2001) poster session},
  volume~68.

\bibitem[\protect\citename{Rehbein \bgroup et al.\egroup
  }2009]{rehbein2009assessing}
Ines Rehbein, Josef Ruppenhofer, and Caroline Sporleder.
\newblock 2009.
\newblock Assessing the benefits of partial automatic pre-labeling for
  frame-semantic annotation.
\newblock In {\em Proceedings of the Third Linguistic Annotation Workshop},
  pages 19--26. Association for Computational Linguistics.

\bibitem[\protect\citename{Rong \bgroup et al.\egroup }2016]{rong2016egoset}
Xin Rong, Zhe Chen, Qiaozhu Mei, and Eytan Adar.
\newblock 2016.
\newblock Egoset: Exploiting word ego-networks and user-generated ontology for
  multifaceted set expansion.
\newblock In {\em Proceedings of the Ninth ACM International Conference on Web
  Search and Data Mining}, pages 645--654. ACM.

\bibitem[\protect\citename{Sadamitsu \bgroup et al.\egroup
  }2012]{sadamitsu2012entity}
Kugatsu Sadamitsu, Kuniko Saito, Kenji Imamura, and Yoshihiro Matsuo.
\newblock 2012.
\newblock Entity set expansion using interactive topic information.
\newblock In {\em PACLIC}, pages 108--116.

\bibitem[\protect\citename{Santorini}1990]{santorini1990part}
Beatrice Santorini.
\newblock 1990.
\newblock Part-of-speech tagging guidelines for the penn treebank project (3rd
  revision).
\newblock {\em Technical Reports (CIS)}, page 570.

\bibitem[\protect\citename{Sarmento \bgroup et al.\egroup
  }2007]{sarmento2007more}
Luis Sarmento, Valentin Jijkuon, Maarten de~Rijke, and Eugenio Oliveira.
\newblock 2007.
\newblock More like these: growing entity classes from seeds.
\newblock In {\em Proceedings of the sixteenth ACM conference on Conference on
  information and knowledge management}, pages 959--962. ACM.

\bibitem[\protect\citename{Settles and Craven}2008]{settles2008analysis}
Burr Settles and Mark Craven.
\newblock 2008.
\newblock An analysis of active learning strategies for sequence labeling
  tasks.
\newblock In {\em Proceedings of the conference on empirical methods in natural
  language processing}, pages 1070--1079. Association for Computational
  Linguistics.

\bibitem[\protect\citename{Shen \bgroup et al.\egroup }2017]{shensetexpan}
Jiaming Shen, Zeqiu Wu, Dongming Lei, Jingbo Shang, Xiang Ren, and Jiawei Han.
\newblock 2017.
\newblock Setexpan: Corpus-based set expansion via context feature selection
  and rank ensemble.
\newblock In {\em The European Conference on Machine Learning and Principles
  and Practice of Knowledge Discovery in Databases (ECML-PKDD 2017)}.

\bibitem[\protect\citename{Tao \bgroup et al.\egroup }2015]{tao2015leveraging}
Fangbo Tao, Bo~Zhao, Ariel Fuxman, Yang Li, and Jiawei Han.
\newblock 2015.
\newblock Leveraging pattern semantics for extracting entities in enterprises.
\newblock In {\em Proceedings of the 24th International Conference on World
  Wide Web}, pages 1078--1088. International World Wide Web Conferences
  Steering Committee.

\bibitem[\protect\citename{Tjong Kim~Sang and
  De~Meulder}2003]{tjong2003introduction}
Erik~F Tjong Kim~Sang and Fien De~Meulder.
\newblock 2003.
\newblock Introduction to the conll-2003 shared task: Language-independent
  named entity recognition.
\newblock In {\em Proceedings of the seventh conference on Natural language
  learning at HLT-NAACL 2003-Volume 4}, pages 142--147. Association for
  Computational Linguistics.

\bibitem[\protect\citename{Tomanek and Hahn}2009]{tomanek2009semi}
Katrin Tomanek and Udo Hahn.
\newblock 2009.
\newblock Semi-supervised active learning for sequence labeling.
\newblock In {\em Proceedings of the Joint Conference of the 47th Annual
  Meeting of the ACL and the 4th International Joint Conference on Natural
  Language Processing of the AFNLP: Volume 2-Volume 2}, pages 1039--1047.
  Association for Computational Linguistics.

\bibitem[\protect\citename{Tsuruoka \bgroup et al.\egroup
  }2008]{tsuruoka2008accelerating}
Yoshimasa Tsuruoka, Jun'ichi Tsujii, and Sophia Ananiadou.
\newblock 2008.
\newblock Accelerating the annotation of sparse named entities by dynamic
  sentence selection.
\newblock {\em BMC bioinformatics}, 9(11):S8.

\bibitem[\protect\citename{Ye \bgroup et al.\egroup }2009]{ye2009conditional}
Nan Ye, Wee~S Lee, Hai~L Chieu, and Dan Wu.
\newblock 2009.
\newblock Conditional random fields with high-order features for sequence
  labeling.
\newblock In {\em Advances in Neural Information Processing Systems}, pages
  2196--2204.

\end{thebibliography}
\bibliographystyle{acl}

\end{document}